\relax
\documentclass[letterpaper]{article} 
\usepackage{aaai21}  
\usepackage{times}  
\usepackage{helvet} 
\usepackage{courier}  
\usepackage[hyphens]{url}  
\usepackage{graphicx} 
\usepackage{natbib}  
\usepackage{caption} 

\usepackage{booktabs}
\usepackage{multirow}
\usepackage{amssymb}

\usepackage{verbatim}
\usepackage{xcolor}
\usepackage{amsmath}

\title{Scribble-Supervised Semantic Segmentation by Random Walk on Neural Representation and Self-Supervision on Neural Eigenspace}
\author{
    Zhiyi Pan\textsuperscript{\rm 1}, Peng Jiang\textsuperscript{\rm 1}\thanks{Corresponding Author }, Changhe Tu\textsuperscript{\rm 1,2}\\
}
\affiliations{

    \textsuperscript{\rm 1}Shandong University \\
    \textsuperscript{\rm 2}AICFVE, Beijing Film Academy\\
    \url{panzhiyi1996@gmail.com}, \url{sdujump@gmail.com}, \url{changhe.tu@gmail.com}

}

\begin{document}

\maketitle

\begin{abstract}
Scribble-supervised semantic segmentation has gained much attention recently for its promising performance without high-quality annotations. Many approaches have been proposed. Typically, they handle this problem to either introduce a well-labeled dataset from another related task, turn to iterative refinement and post-processing with the graphical model, or manipulate the scribble label. This work aims to achieve semantic segmentation supervised by scribble label directly without auxiliary information and other intermediate manipulation. Specifically, we impose diffusion on neural representation by random walk and consistency on neural eigenspace by self-supervision, which forces the neural network to produce dense and consistent predictions over the whole dataset. The random walk embedded in the network will compute a probabilistic transition matrix, with which the neural representation diffused to be uniform. Moreover, given the probabilistic transition matrix, we apply the self-supervision on its eigenspace for consistency in the image's main parts. In addition to comparing the common scribble dataset, we also conduct experiments on the modified datasets that randomly shrink and even drop the scribbles on image objects. The results demonstrate the superiority of the proposed method and are even comparable to some full-label supervised ones. The code and datasets are available at~\url{https://github.com/panzhiyi/RW-SS}.



\end{abstract}

\section{Introduction}

\begin{figure*}[t]
\centering
\includegraphics[width=0.9\textwidth]{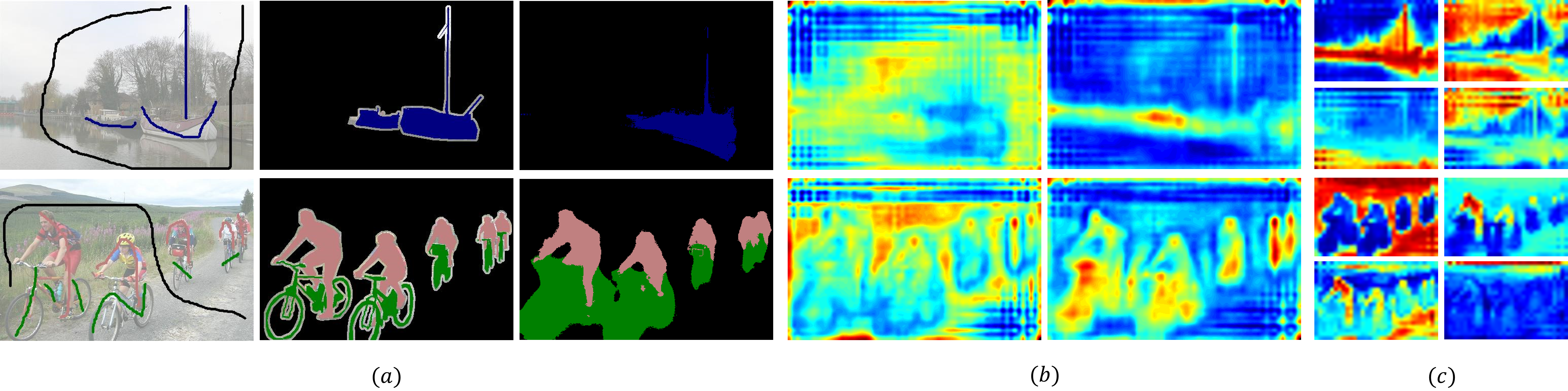}
\caption{Results and Intermediate Visualizations. (a) From left to right: scribble-annotation, ground-truth, and our prediction. (b) From left to right: Neural representation before and after the random walk. (c) Leading eigenvectors of the transition matrix.}
\label{fig:main}
\end{figure*}

In recent years, the use of neural networks, especially convolutional neural networks, has dramatically improved semantic classification, detection, and segmentation~\cite{he2016deep,long2015fully,wang2018non}. As one of the most fine-grained ways to understand the scene, typically, semantic segmentation demands large-scale data with high-quality annotations to feed the network. However, the pixel-level annotating process for semantic segmentation is costly and tedious, limiting its flexibility and usability on some tasks that require rapid deployment~\cite{lin2016scribblesup}. As a consequence, the scribble annotations, which are more easily available, become popular.

Correspondingly, some datasets and approaches are developed. \citealp{lin2016scribblesup} proposed a scribble-annotated dataset based on the pascal VOC~\cite{everingham2015pascal} and adopted the classic graphical model as post-processing to obtain the final dense predictions. \citealp{vernaza2017learning} and \citealp{wang2019boundary} optimize the performance of semantic segmentation by introducing an auxiliary task, edge detection. However, this edge detection task needs to be trained on another well-labeled dataset, so these methods have not relieved the heavy labor of annotation yet.
To avoid the post-processing and dependence on another well-labeled dataset, \citealp{tang2018normalized} and \citealp{tang2018regularized} design graphical model based regularized loss to make the predictions between similar pixels consistent. But these two works only measured similarities in color and texture, did not consider semantic similarity for regularization. Moreover, most of these methods require every existing object in the image labeled, which is too strict for dataset preparation. In this work, we intend to propose a more flexible approach to get rid of the above issues holistically.

The proposed approach tackles scribble-supervised semantic segmentation by leveraging the uniform and consistency of neural representation (features). A representative result is shown in Fig.~\ref{fig:main}(a). The assumption is that, given the sparse scribble labels, if we guide the neural representation to be uniform within each image's objects and consistent between related images, then the neural network's best solution to the cross-entropy loss is the dense semantic predictions. To realize this kind of learning trend, we impose the diffusion on neural representation by random walk and consistency on neural eigenspace by self-supervision during training.

The random walk on neural representation has been studied in some works and could produce uniform and dense semantic predictions within each object~\cite{jiang2018difnet}. The key to this kind of random walk is forming a probabilistic transition matrix that measures the similarity between neural representations. 
Then, with the transition matrix, the neural representation is diffused to be uniform. 
Fig.~\ref{fig:main}(b) shows several neural representations before and after the random walk.
In addition to the uniform neural representation within each object, it is also essential to have consistent neural representation over related images to produce dense and confident semantic predictions. Namely, when the image is transformed, the neural representation should be the same as the original one with the corresponding transform. This kind of consistency is usually referred to as self-supervision~\cite{laine2016temporal,tarvainen2017mean,mittal2019semi} and measured on neural representation. In this way, though with the various and sparse scribble labels, the network will still tend to generate consistent object perception. 

However, for semantic segmentation, the consistency over the whole image is not necessary. When some parts of the image are distorted and changed heavily after transform, it is hard for the network to generate consistent neural representation anymore and may confuse the network in some scenarios. In this work, we propose to set the self-supervision on the main parts of images by imposing the consistent loss on the eigenspace of the transition matrix. The idea is inspired by spectral methods~\cite{von2007tutorial}, which observed that the eigenvectors of the Laplacian matrix have the capability to distinguish the main parts in the images, and some methods use this property for clustering~\cite{ng2002spectral,nadler2007fundamental} and saliency detection~\cite{jiang2019super,yang2013saliency}. Since the eigenspace of the transition matrix has a close relation to the one of the Laplacian matrix, our self-supervision on the transition matrix's eigenspace will also focus on the main image parts. Several leading eigenvectors are presented in Fig.~\ref{fig:main}(c).


The computation of eigenspace is time-consuming and unstable, especially during the dynamic optimizing of the neural network. Though some people have developed approximation methods~\cite{dang2018eigendecomposition,wang2019backpropagation,sun2019neural}, it is better to avoid the explicit eigenspace decomposition. Thus, in our implementation, we only apply a soft consistent loss on the eigenspace. For eigenvalue consistency, according to the fact that the matrix's trace is equal to the sum of its eigenvalues, we measure the matrix trace consistency instead. Given the consistency on eigenvalue, we compute the Kullback-Leibler Divergence between probabilistic transition matrices to further prompt eigenvectors' consistency. We also developed convenient ways to compute consistent loss regarding the complicated relationship between probabilistic transition matrices after image transform and modification.

The proposed method demonstrates consistent superiority to others on the common scribble-annotated dataset and is even comparable to some fully supervised ones. Moreover, we further conducted experiments when the scribble-annotations gradually shrank and dropped. The proposed method could still work reasonably, even the scribble shrank to the point or dropped significantly. Besides, careful ablation study and mechanism study is made to verify the effectiveness of every module. Finally, the code and dataset are open-sourced.

\section{Related Work}

\subsection{Scribble-Supervised Semantic Segmentation}
The scribble-supervised semantic segmentation aims to produce dense predictions given only sparse scribble-annotations. Existing deep learning based works usually could be divided into two groups: 1) Two-stage approaches~\cite{lin2016scribblesup,vernaza2017learning}, which firstly obtain the full mask pseudo-labels by manipulating the scribble annotations, then train the semantic segmentation network as usual with pseudo-labels.
2) Single-stage approaches~\cite{tang2018normalized,tang2018regularized}, which directly train the network using scribble-annotations by the specific design of loss function and network structure.
While two-stage approaches can be formulated as regular semantic segmentation, single-stage approaches are usually defined to minimize $L$:
\begin{equation}
    L=\sum_{p\in\Omega_{\mathcal{L}}}c(s_p,y_p)+\lambda\sum_{p,q\in\Omega}u(s_p,s_q),
\label{eq:eq1}
\end{equation}
where $\Omega$ is the pixel set, $\Omega_{\mathcal{L}}$ is the pixel set with scribble-annotations, $s_i$ represents the prediction of pixel $i$, and $y_i$ is the corresponding ground truth. The first term measures the error with respect to the scribble annotations and usually is in the form of cross-entropy. The second term is a pair-wise regularisation to help generate uniform prediction. The two terms are harmonized by a weight parameter $\lambda$.

For scribble-supervised semantic segmentation, the graphical model has been prevalently adopted in either two-stage approaches for generating pseudo-label or one-stage approaches for loss design. \citealt{lin2016scribblesup} iteratively conduct label refinement and network optimization through a graphical model. \citealt{vernaza2017learning} generate high-quality pseudo-labels for full-label supervised semantic segmentation by optimizing graphical model with edge detector learned from another well-labeled dataset. These two works require iterative optimization or auxiliary dataset. Instead, \citealt{tang2018normalized, tang2018regularized} add the soft graphical model regularization into loss function and avoid explicitly graphical model optimization. Besides, some works only well on the dataset with every existing object labeled by at least one scribble. In general, most methods have not provided a flexible and efficient solution to scribble supervised semantic segmentation yet.
\subsection{Random Walk}
Uniform neural representation is crucial for semantic segmentation to produce dense predictions no matter scribble supervised or full-label supervised. Typically, embedding a random walk operation in the network would provide help. In this way, \citeauthor{bertasius2017convolutional} use a random walk to addresses the issues of poor boundary localization and spatially fragmented predictions. Then \citeauthor{jiang2018difnet} further conduct a sequence of random walks to approximate a stationary of the diffusion process. A random walk operation in the network can be defined as:
\begin{equation}
    f(x)^{L}=\alpha{Pf(x)^{L-1}}+f(x)^{L-1},
\label{eq:eq2}
\end{equation}
where $f(x)^{L-1}$ is the neural representation of image $x$ in layer $L$-$1$ and $f(x)^{L}$ is the neural representation after random walk in layer $L$. $\alpha$ is the weight parameter learned during training. The most key component of random walk is the probabilistic transition matrix $P$, whose unit $p_{ij}$ measures the similarity between $i$-th and $j$-th elements of neural representations. Besides, all the units are positive, with every row of the matrix is summed to $1$. Inner product, embedded gaussian~\cite{wang2018non}, and diffusion distance~\cite{sun2019neural} have been widely used to compute the similarity between neural representations.

\subsection{Self-Supervision}
Consistency on neural representation is the property that should be concerned in almost all the learning-based tasks. Consistency would be helpful for detection, tracking, and definitely for segmentation. For this property, the consistent loss has been widely adopted.
Since no ground truth required, this loss is more popular for unsupervised and semi-supervised tasks, and usually defined as the difference between neural representations of image and its transform:
\begin{equation}
    ss(x,\phi)=l(T_\phi(f(x)),f(t_\phi(x))),
\label{eq:eq3}
\end{equation}
where $t_\phi$ denotes the transform operation on $x$ with $\phi$ parameter, while $T_\phi$ corresponds to the transform operation on $f(x)$ ($t_\phi$ and $T_\phi$ are pair of corresponding transforms for self-supervision). $l$ is the metric to define how the difference is measured. This kind of consistent loss is also referred to as self-supervision, which we denote as $ss(x,\phi)$. Self-supervision usually has to conduct two feed-forward processes, \citealp{laine2016temporal} propose temporal ensembling, which saves previous neural representations to avoid multiple feed-forward and facilitate the computation. Then, the mean teacher~\cite{tarvainen2017mean} proposed to prepare a teacher network rather than save the auxiliary information. The self-supervision has been used for semi-supervised semantic segmentation~\cite{mittal2019semi}.

\section{Method}

\begin{figure*}[t]
\centering
\includegraphics[width=0.8\textwidth]{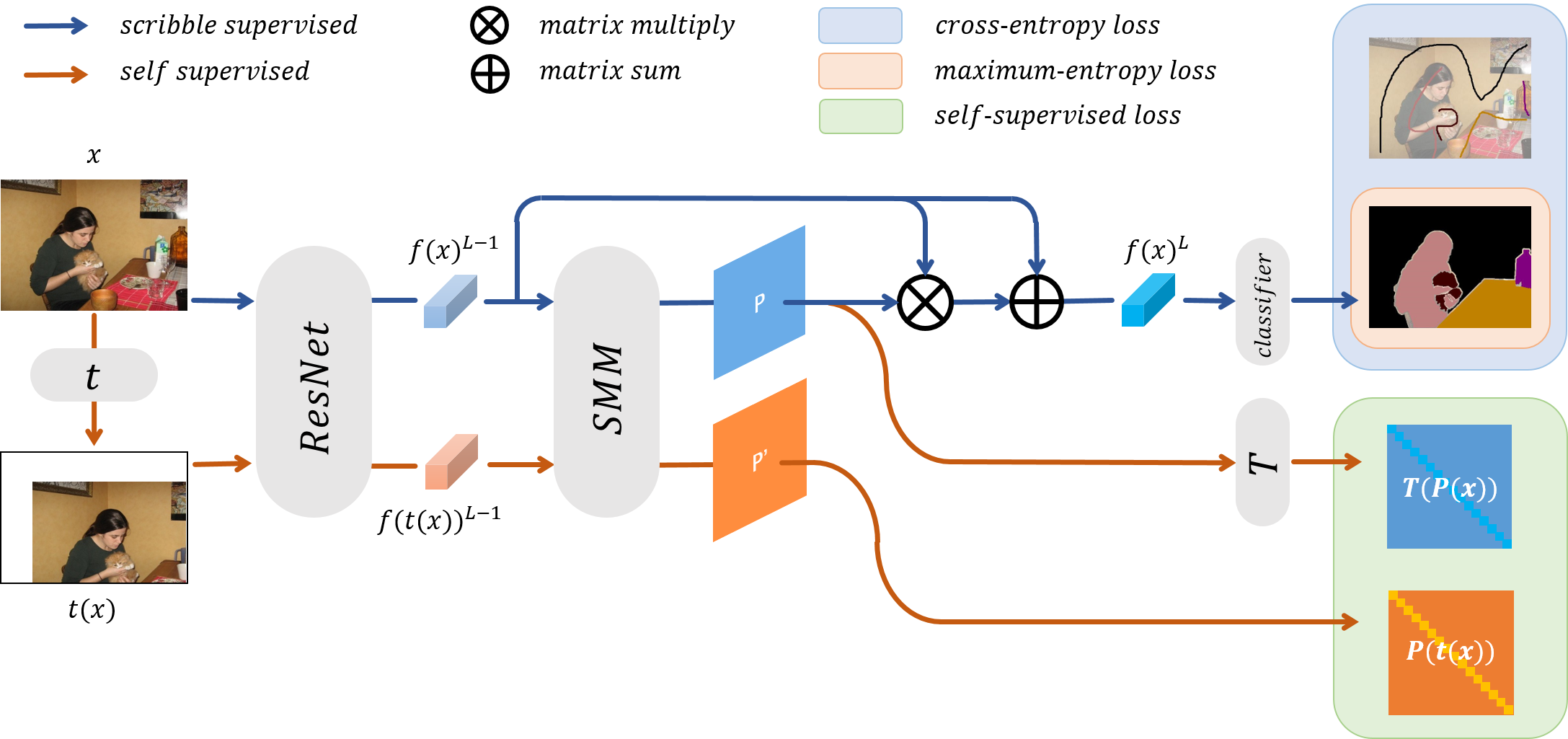}
\caption{Network Architecture and Pipeline. We use blue flow to represent the scribble-supervised training and orange flow to represent self-supervised training. Given an image and its transform, we pass them to ResNet backbone to extract neural representations, from which the similarity measurement module (SMM) computes transition matrices. Then, a random walk is carried out on the neural representation of the original image. The results are used for classifying semantic. Simultaneously, the self-supervised loss is set between the transition matrices to realize the self-supervision on neural eigenspace. During inference, only blue flow is activated.}
\label{fig:pipeline}
\end{figure*}

The network of the proposed method is illustrated in Fig.~\ref{fig:pipeline}, including two modules (ResNet backbone to extract features, and Similarity Measurement Module to compute the probabilistic transition matrix), one specific process (the random walk process) and three loss functions (the common cross-entropy loss, the maximum-entropy loss and self-supervised loss). In the following subsections, we will introduce the main components and discuss the insight.

\subsection{Similarity Measurement Module}
Similarity Measurement Module computes the distance between any pairs of neural representation elements and forms the probabilistic transition matrix. The module is illustrated in Fig.~\ref{fig:pipeline}. There are several choices for the distance definition, such as Euclidean distance, inner product~\cite{wang2018non}, and diffusion distance~\cite{sun2019neural}. To preserve the property of the transition matrix (positive unit and each row sums to 1) and compute efficiency, in this work, we use the inner product with softmax operation to form the transition matrix. The probabilistic transition matrix $P$ is defined as:
\begin{equation}
    P=softmax({f(x)^{L-1}}^Tf(x)^{L-1}),
\label{eq:eq4}
\end{equation}
where $f(x)^{L-1}$ is the neural representation of input $x$ in layer $L$-1 and has flatted to $MN$$\times$$C$ dimension ($M$: height, $N$: width, $C$: feature channels). Thus, ${f(x)^{L-1}}^Tf(x)^{L-1}$ will produce a matrix of dimension $MN$$\times$$MN$. With $softmax$ in the horizontal direction, we generate the adequate probabilistic transition matrix $P$.


\subsection{Embedded Random Walk}
We embed the random walk process in the computation flow of the neural network. The process is defined as Eq.~\ref{eq:eq2} (with $P$ as Eq.~\ref{eq:eq4}) and learned $\alpha$ to control the degree of random walk, and conducted on the final layer right before the classifier.

Through the random walk process, the $i$-th element in the neural representation of layer $L$, $f(x)_i^L$, equals to the weighted sum of all the elements in layer $L$-$1$ with the weight defined by the $i$-th row of $P$:
\begin{equation}
    f(x)_i^L=\alpha\sum_{j=1}^{MN} P_{ij}f(x)_{j}^{L-1}+f(x)_i^L.
\label{eq:eq5}
\end{equation}
More similar $f(x)_j^{L-1}$ to $f(x)_i^{L-1}$, more large the $P_{ij}$ is, and more important $f(x)_j^{L-1}$ to the $f(x)_i^L$.

With this process, each element in neural representation have relationships with all the other elements. Thus, the scribble annotations will affect not only the labeled elements but also unlabeled elements. When the cross-entropy loss is applied, the best solution (achieve lower loss) will be the uniform predictions within each object, considering the random walk process's constraint. In Fig.~\ref{fig:main}(b), we visualize the neural representations (after the sum of absolute values in the channel dimension) before and after the random walk process. As can be seen, the neural representations become uniform within each semantic region after the random walk. It verifies our assumption in the introduction. It is worth noting that we have not imposed supervision on $P$, but $P$ has gained semantic similarity knowledge from the results.
It is the embedded random walk process with the scribble-annotations guide the formulation of $P$ to produce uniform predictions. 

\subsection{Self-Supervision Loss}
The self-supervision loss computes the difference between the neural representations of the image and its transform. There are several issues that need to be considered when applying self-supervision loss in this work: (1) where the self-supervision is involved; (2) how the self-supervision loss is calculated; (3) what kinds of the transform will be used. We address these issues in the following.

\subsubsection{Self-Supervision on Eigenspace}
In our work, the typical choice of neural representation for self-supervision is $f(x)^L$, and thus the self-supervision loss will be
\begin{equation}
    ss(x,\phi)=l(T_\phi(f(x)^L),f(t_\phi(x))^L).
\label{eq:eq6}
\end{equation}
However, as for semantic segmentation tasks with self-supervision, we argue that directly calculating loss on the whole neural representation is not necessary and may not optimal. When the image is distorted heavily after the transform, some parts of its neural representation will change greatly, so minimizing Eq.~\ref{eq:eq6} will be hard and even ambiguous.

The transition matrix $P$ could also be defined as $P=D^{-1}W$, where $W$ is the affinity matrix, and $D$ is the degree matrix. The eigenspace of $P$ and the one of normalized Laplacian matrix $L$ have close relationships, considering $L=D^{-1}(D-W)$. It can be proved that $\Lambda_P$=$1-\Lambda_L$ and $U_P$=$U_L$ ($\Lambda$ denotes diagonal matrix with eigenvalues as entries, $U$ denotes matrix with eigenvectors as columns). According to \citealp{von2007tutorial,Jiang_2015_ICCV,jiang2019super}, columns of $U_L$ have the capability to distinguish the main parts of the images. So, $U_P$ will also inherit this property. 

We visualize several eigenvectors of $P$ in Fig.~\ref{fig:main}(c). As can be seen, compared with original neural representation, the eigenvectors of $P$ are more powerful to distinguish the main parts from others and neglect some details, though $P$ is also computed from neural representation. Based on the above analysis, in this work, we propose to set the self-supervision on the eigenspace of $P$:
\begin{equation}
\begin{aligned}
    ss(x,\phi)&=l(T_\phi(U_P(x)),U_P(t_\phi(x)))\\
    &+l(T_\phi(\Lambda_P(x)),\Lambda_P(t_\phi(x))).
\end{aligned}
\label{eq:eq7}
\end{equation}

\subsubsection{Soft Eigenspace Self-Supervision}
Eq.~\ref{eq:eq7} requires explicit eigendecomposition, which is time-consuming, especially within the deep neural network context. Though there are some approximation methods~\cite{dang2018eigendecomposition,wang2019backpropagation,sun2019neural} proposed, their efficiency and stability are still far from satisfactory. To this end, we develop soft eigenspace self-supervision, which avoids explicit eigendecomposition. Firstly, in view of the fact that the matrix's trace is equal to the sum of its eigenvalues, we measure the consistency on the $\Lambda$ by computing the difference on the trace of $P$, $tr(P)$. Secondly, given the consistency on the $\Lambda$, we propose to measure the consistency on the $P$ to obtain consistent $U$ indirectly. In 
other words, the soft eigenspace self-supervision loss is defined as:
\begin{equation}
\begin{aligned}
    ss_P(x,\phi)&=l_1(T_\phi(P(x)),P(t_\phi(x)))\\
    &+\gamma\ast l_2(T_\phi(tr(P(x))),tr(P(t_\phi(x)))),
\end{aligned}
\label{eq:eq8}
\end{equation}
where $P(x)$ denotes $P$ for image $x$, $tr(P(x))$ is the trace of $P(x)$. Since $P(x)$ is the probabilistic transition matrix, we use Kullback-Leibler Divergence as $l_1$ to measure the difference. $l_2$ is defined as $L_2$ norm. $\gamma$ is the weight to control the two terms.

\subsubsection{Transform Operation and Computing Matrix}
\begin{figure}[t]
\centering
\includegraphics[width=0.4\textwidth]{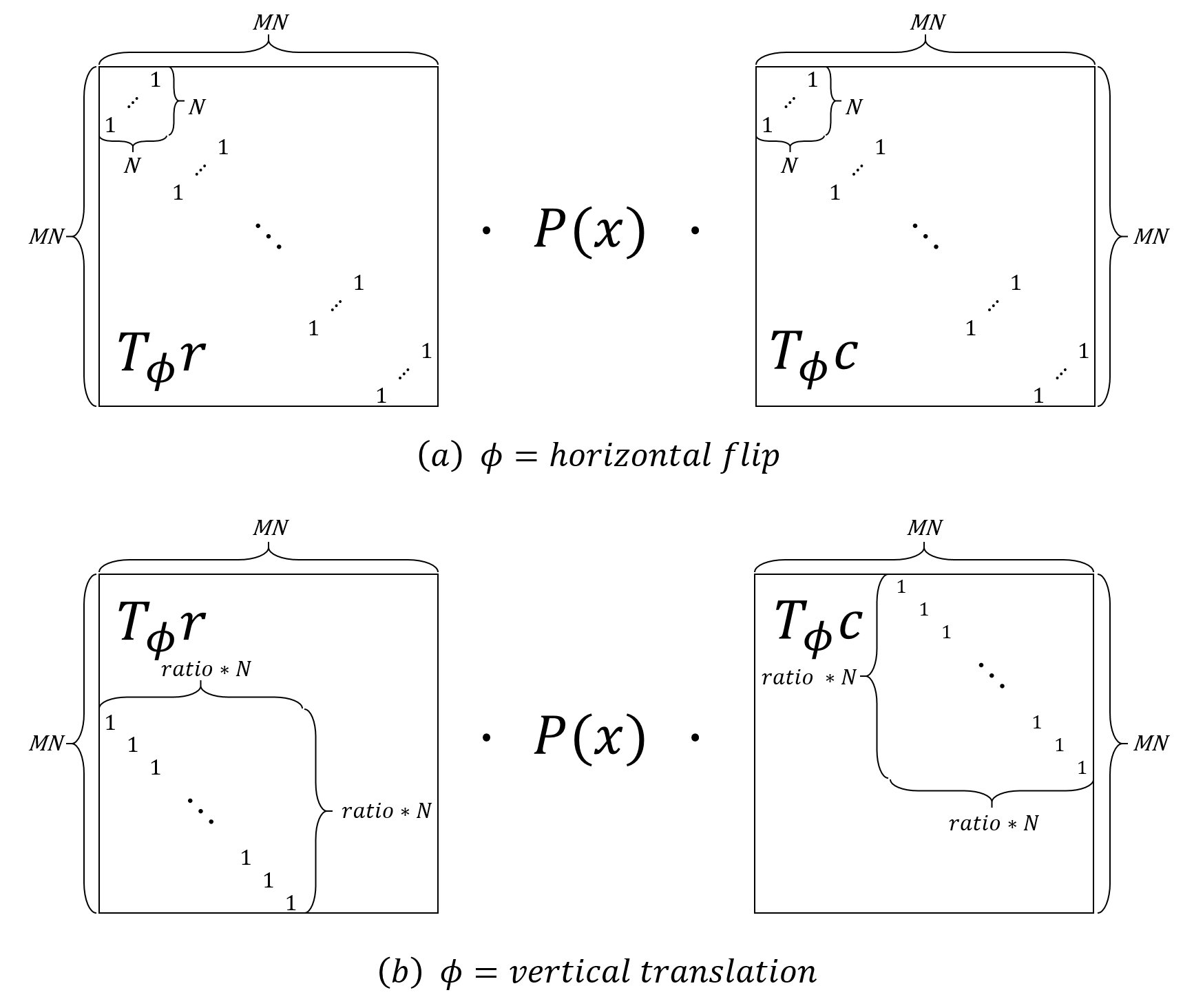}
\caption{Visualization of predefined computing matrices for self-supervision on eigenspace.}
\label{fig:ss_op}
\end{figure}

In this work, we have linear transforms, including horizontal flip and translation, $\phi\in$(horizontal flip, translation). Compared with the transform effect on the neural representation, any transform will lead to a complex change on $P$ and complicate the computation. However, since all the transform is linear, the probabilistic transition matrix after transform can be expressed as the multiplication of the original $P$ with the predefined computing matrices, to facilitate the Eq.~\ref{eq:eq8} computation. $T_{\phi}(P(x))$ can be defined as:
\begin{equation}
\begin{aligned}
    T_{\phi}(P(x))=T_{\phi}r\cdot{P(x)}\cdot{T_{\phi}c},
\end{aligned}
\end{equation}
where $T_{\phi}r$ and $T_{\phi}c$ are predefined computing matrices for transform $\phi$. In Fig.~\ref{fig:ss_op}, we visualize computing matrices for horizontal flip and vertical translation when using soft eigenspace self-supervision. Please check supplementary for the detail definitions.

\subsection{Maximum-entropy Loss}
To force the network producing a more convinced prediction, we further minimize the maximum-entropy loss on the final prediction, which is defined as
\begin{equation}
\begin{aligned}
    E(s)=-\frac{1}{HW} \sum_{i,j}\sum_c( s(i,j,c)\cdot log(s(i,j,c))) ,
\end{aligned}
\end{equation}
where $s$ represents the final prediction, and is of the size $H\times W\times C$ ($C$ is the number of categories). $s(i,j,c)$ represents the probability that the pixel at position $(i,j)$ of the image belongs to the $c$-th category.

\begin{table}
    \centering
    \setlength{\belowcaptionskip}{3pt}
    \caption{Ablation study on random walk, and operation and location of self-supervision.}
    \begin{tabular}{c|c|c|c}
    \toprule[1pt]
        \multirow{2}{*}{Random Walk}&\multicolumn{2}{|c|}{Self-Supervision}&\multirow{2}{*}{mIoU}\\
        \cline{2-3}
        ~&operation&location&~\\
        \hline
        \hline
        &-&-&64.4\\
        
        $\checkmark$&-&-&67.6\\
        \hline
        $\checkmark$&flip&$f(x)^{L-1}$&69.8\\

        $\checkmark$&flip&$f(x)^L$&70.1\\

        $\checkmark$&flip&$Eigenspace$&\textbf{70.5}\\
        \hline
        $\checkmark$&translation&$f(x)^{L-1}$&70.5\\

        $\checkmark$&translation&$f(x)^L$&70.5\\
        
        $\checkmark$&translation&$Eigenspace$&\textbf{70.8}\\
        \hline
        $\checkmark$&random&$f(x)^{L-1}$&70.2\\

        $\checkmark$&random&$f(x)^L$&70.3\\
        
        $\checkmark$&random&$Eigenspace$&\textbf{71.2}\\
    \bottomrule[1pt]
    \end{tabular}
    \label{tab:ablation}
\end{table}

\section{Experiment}
\subsection{Implementation}
The whole pipeline is shown in Fig.~\ref{fig:pipeline}. We use pre-trained ResNet~\cite{he2016deep} with dilation~\cite{chen2017deeplab} as the backbone to extract initial neural representations. The total loss in our work is defined as:
\begin{equation}
\begin{aligned}
L=\sum_{p\in\Omega_{\mathcal{L}}}c(s_p,y_p)+\omega_1\ast E(s)+\omega_2\ast ss_P(x,\phi),
\end{aligned}
\label{eq:eq9}
\end{equation}
where $\omega_1$ and $\omega_2$ are the predefined weights. All the training images are randomly scaled (0.5 to 2), rotated (-10 to 10), blurred, and flipped for data augmentation, then cropped to $465 \times 465$ before feeding to the network. And the immediate output of $ResNet$ ($f(x)^{L-1}$) is of spatial dimension $29 \times 29$. The training process has two steps. Firstly, only the cross-entropy loss is used to train the network because the network may not perform well initially, and in this time, self-supervision will not bring benefits but prevent the optimization. After the network got reasonable performance, the whole Eq.~\ref{eq:eq9} (scribble supervise and self supervise) is activated. The process has been visualized in Fig.~\ref{fig:pipeline}. The two steps training is also used in~\cite{tang2018normalized, tang2018regularized}.

\begin{figure*}[t]
\centering
\includegraphics[width=0.8\linewidth]{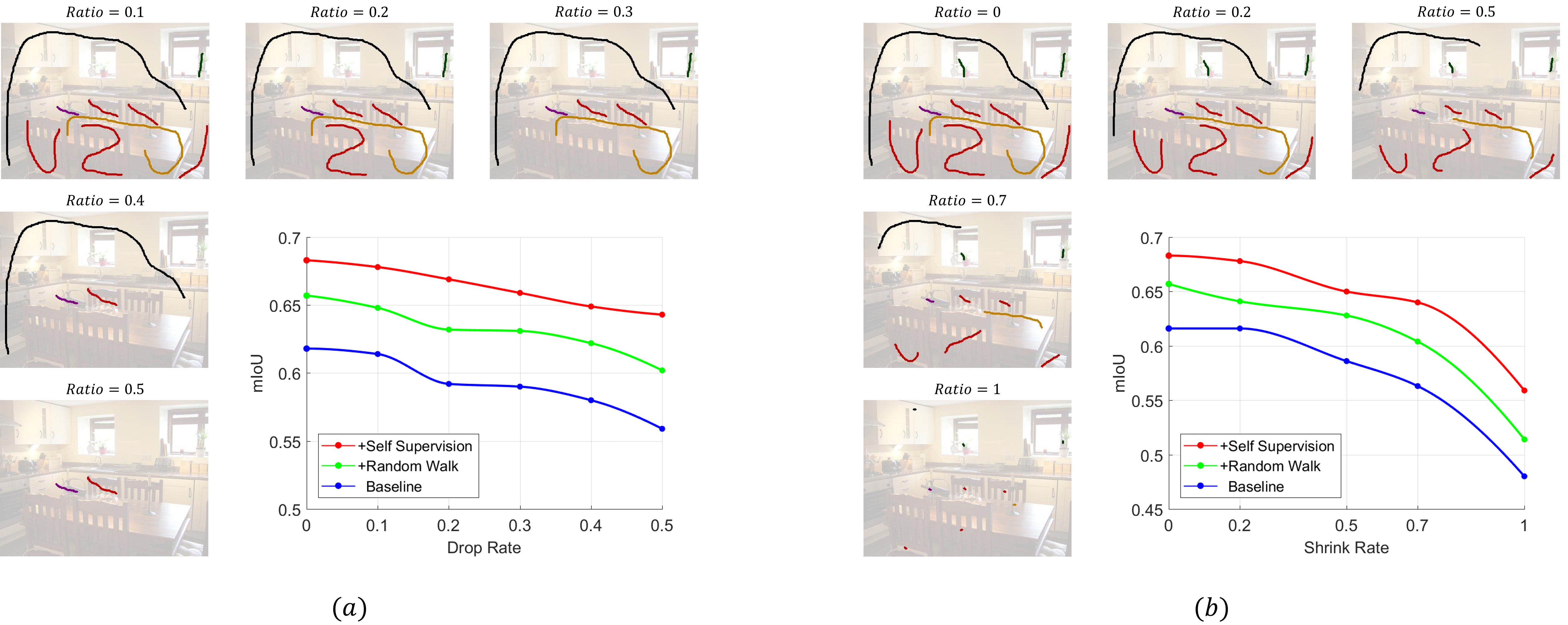}
\caption{(a) A representative sample of \emph{scribble-drop} with the scribble drop rate from 0.1 to 0.5, and the mIoU scores on different settings.
(b) A representative sample of \emph{scribble-shrink} with the scribble shrink rate from 0 to 1 (point), and the mIoU scores on different settings.
(Zoom in for better visualization)}
\label{fig:datasets}
\end{figure*}

\begin{table}
    \centering
    \setlength{\belowcaptionskip}{3pt}
    \caption{Performance on the validation set of pascal VOC. The supervision types (Sup.) indicate: $\mathcal{P}$–point, $\mathcal{S}$–scribble, $\mathcal{B}$–bounding box, $\mathcal{I}$–image-level label, and $\mathcal{F}$–full label.}
    \scalebox{0.9}{
    \begin{tabular}{l|c|c|c|c}
    \toprule[1pt]
        Method&Sup.&Backbone&wo/ CRF&w/ CRF\\
        \hline
        \hline
        What'sPoint&$\mathcal{P}$&$VGG16$&46.0&-\\
        
        SDI&$\mathcal{B}$&$ResNet101$&-&69.4\\
        
        BCM&$\mathcal{B}$&$ResNet101$&-&70.2\\
        
        CIAN&$\mathcal{I}$&$ResNet101$&64.1&67.3\\
        
        FickleNet&$\mathcal{I}$&$ResNet101$&64.9&-\\
        
        SCE&$\mathcal{I}$&$ResNet101$&64.8&66.1\\

        DeepLabV2&$\mathcal{F}$&$ResNet101$&76.4&77.7\\
        \hline
        scribblesup&$\mathcal{S}$&$VGG16$&-&63.1\\

        RAWKS&$\mathcal{S}$&$ResNet101$&59.5&61.4\\
        
        NCL&$\mathcal{S}$&$ResNet101$&72.8&74.5\\

        KCL&$\mathcal{S}$&$ResNet101$&73.0&75.0\\
        
        BPG-PRN&$\mathcal{S}$&$ResNet101$&71.4&-\\
        \hline
        
        ours-ResNet50&$\mathcal{S}$&$ResNet50$&71.9&73.6\\
        

        ours-ResNet101&$\mathcal{S}$&$ResNet101$&\textbf{73.4}&\textbf{75.8}\\
    \bottomrule[1pt]
    \end{tabular}
    \label{tab:all}}
\end{table}

\subsection{Experiment setting}
\subsubsection{Datasets}\label{sec:dataset}
We mainly compare with others on the common scribble-annotated dataset, \emph{scribblesup} ~\cite{lin2016scribblesup}. This dataset is revised from pascal VOC~\cite{everingham2015pascal} and has every existing object in each image labeled by at least one scribble. However, our method does not need this hypothesis. To better demonstrate the advantage of the proposed method, we further proposed two variants of \emph{scribblesup}. The first one is \emph{scribble-drop}, where every object in images may drop (\emph{i.e.} delete) all the scribble annotations with a defined possibility. The second one is \emph{scribble-shrink}, where every scribble in the image is shrunk randomly (even to a point). All the images in two datasets are identical to the \emph{scribblesup}'s, as well as the training and validation partition. In our experiments, many settings of drop and shrink rate are tested. Fig.~\ref{fig:datasets} shows several representative samples of \emph{scribble-drop} and \emph{scribble-shrink}.

\subsubsection{Compared methods}
We compare with recently proposed scribble-supervised methods including scribblesup~\cite{lin2016scribblesup}, RAWKS~\cite{vernaza2017learning}, NCL~\cite{tang2018normalized}, KCL~\cite{tang2018regularized} and BPG-PRN~\cite{wang2019boundary}, and also other weakly-supervised methods such as point supervised (What’sPoint~\cite{bearman2016s}), bounding-box supervised (SDI~\cite{khoreva2017simple}, BCM~\cite{Song_2019_CVPR}) and image-level-label supervised (CIAN~\cite{fan2020cian}, FickleNet~\cite{lee2019ficklenet}, SCE~\cite{chang2020weakly}).
Besides, full-label supervised method (DeepLabV2~\cite{chen2017deeplab}) is also compared. We use $mIoU$ as the main metric to evaluate these methods and ours. When comparing with others, we mainly refer their reported scores if available.

\subsubsection{Hyper-parameters}
The proposed method has 200 epochs of training, with the first 100 epochs have no self-supervised loss. For every step, eight images (batch size) are randomly selected to train the network with Adam~\cite{kingma2014adam} optimizer, Sync-BatchNorm~\cite{ioffe2015batch} and learning rate as $1e$-$3$ for the first 100 epochs and $1e$-$4$ for the rest. The weights $\gamma$, $\omega_1$ and $\omega_2$ are set to be 0.01, 0.2 and 1, respectively. Besides, as the common way for semantic segmentation~\cite{zhao2018psanet}, data augmentation is adopted during training. All the computations are carried out on NVIDIA TITAN RTX GPUs.

\begin{figure}[t]
\centering
\includegraphics[width=0.45\textwidth]{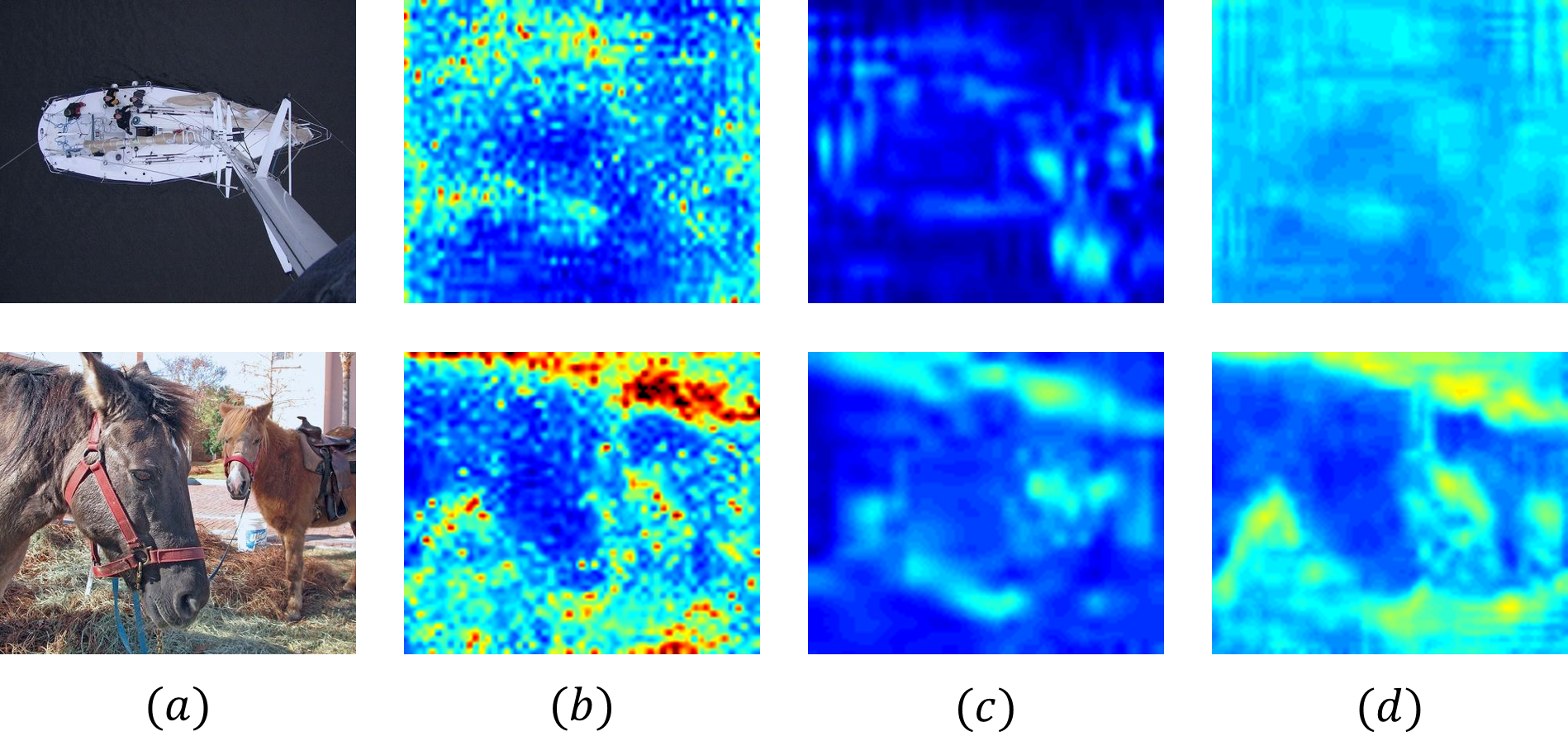}
\caption{Visualization of variation by different self-supervision operations. (a) The input image, (b) The variation on $f(x)^{L-1}$, (c) The variation on $P(x)$, (d) The variation on $f(x)^L$. The first row shows the variation for the flip operation, while the second row is for the translation operation.}
\label{fig:variations1}
\end{figure}

\begin{figure*}[htbp]
\centering
\includegraphics[width=1\linewidth]{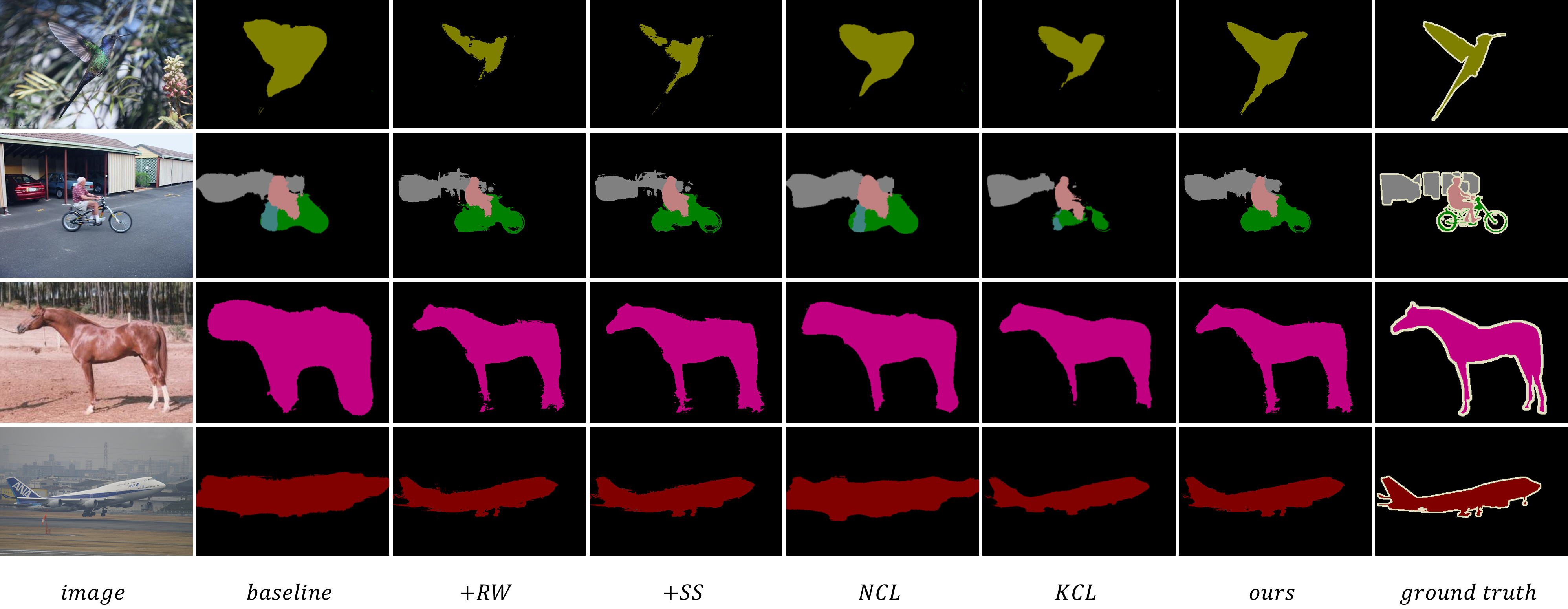}
\caption{Visual comparison between proposed method and others on \emph{scribblesup} dataset.}
\label{fig:VisualizationResults}
\end{figure*}

\begin{table}
    \centering
    \setlength{\belowcaptionskip}{3pt}
    \caption{Variation comparison under the same transform.}
    \begin{tabular}{c|c|c|c}
    \toprule[1pt]
    &$f(x)^{L-1}$&$f(x)^L$&$P(x)$\\
    \hline
    \hline
    flip&25.3\%&27.7\%&11.1\%\\
    
    translation&7.7\%&15.2\%&6.9\%\\
    \bottomrule[1pt]
    \end{tabular}
    \label{tab:ablation2}
\end{table}

\begin{table*}
    \centering
    \setlength{\belowcaptionskip}{3pt}
    \caption{The performance drop ratios compared to no drop and no shrink when only using baseline and gradually adding random walk and self-supervision.}
    \begin{tabular}{l|c|c|c|c|c||l|c|c|c|c}
    \toprule[1pt]
    drop rate&0.1&0.2&0.3&0.4&0.5&shrink rate&0.2&0.5&0.7&1\\
    \hline
    \hline
    baseline&0.7\%&4.2\%&4.5\%&6.2\%&9.6\%&baseline&0\%&4.9\%&8.6\%&22.1\%\\
    
    +Random Walk&1.4\%&3.8\%&4.0\%&5.3\%&8.4\%&+Random Walk&2.4\%&4.4\%&8.1\%&21.8\%\\
    
    +Self-Supervision&0.7\%&2.1\%&3.5\%&5.0\%&5.9\%&+Self-Supervision&0.7\%&4.8\%&6.3\%&18.2\%\\
    \bottomrule[1pt]
    \end{tabular}
    \label{tab:drop_ratios}
\end{table*}

\subsection{Ablation Study}
In Tab.~\ref{tab:ablation}, we do ablation study w/wo random walk, w/wo self-supervision, and self-supervision on $f(x)^{L-1}$, $f(x)^{L}$ and $P(x)$ on \emph{Scribblesup} dataset. We use \emph{ResNet50} as the backbone without maximum-entropy loss and CRF to study the random walk and self-supervision more thoroughly. The first one is a fully convolutional neural network (Baseline). With random walk, the mIoU receives more than 3\% promotion. After incorporating self-supervision (the last one), the performance further boosted by 3.6\% increasing. As for the self-supervision operations, we observe that both are helpful and self-supervision on the eigenspace outperforms on the other locations consistently no matter what kind of operation applied. Besides, randomly combined on the eigenspace leads to the best performance, while other locations do not show such promotion. It is worth noting that eigenspace is also a kind of neural representation that lies in the middle between $f(x)^{L-1}$ and $f(x)^L$. The middle one always outperforms sides, indicating a general advantage of self-supervision on neural eigenspace. 

In Tab.~\ref{tab:ablation2}, we show the mean variations of $f(x)^{L-1}$, $f(x)^L$ and $P(x)$ under the same transform (no self-supervision applied yet). The variation is measured by the relative error defined as $|T_\phi(f)-f'|/(|T_\phi(f)|+|f'|)$ ($f$: feature, $f'$: feature after transform). As can be seen, the same transform will always lead to less change on $P(x)$. Considering the property of eigenvectors, we believe $P(x)$ is not sensitive to the trivial structures (whose semantic may be heavily changed after transform). In Fig.~\ref{fig:variations1}, we visualize the variation of two images by different self-supervision operations, respectively. It can be seen the variation on $P(x)$ are most distributed on the main objects or object boundaries, while $f(x)^{L-1}$ and $f(x)^L$ also highlight backgrounds with uneven distribution. This phenomenon indicates that self-supervision on $P(x)$ will be more effortless and less confusing.


\subsection{Quantitative Results}
After introducing maximum-entropy loss, our method gets $71.9\%$ mIoU with $ResNet50$ and $73.4\%$ with $ResNet101$ on $Scribblesup$ dataset. When comparing with others, we also report the performance with CRF as others. Tab.~\ref{tab:all} lists all the scores of compared methods under different settings. In addition to the scribble-supervised methods, we also show methods with other type labels, including point, bounding-box, image-level-label and full-label. The proposed method reaches state-of-the-art performance compared with scribble and other weakly supervised methods and is even comparable to the full-label one. The reported full-label method (DeepLabV2) had been pre-trained on COCO dataset.

\begin{figure}[htbp]
\centering
\includegraphics[width=1\linewidth]{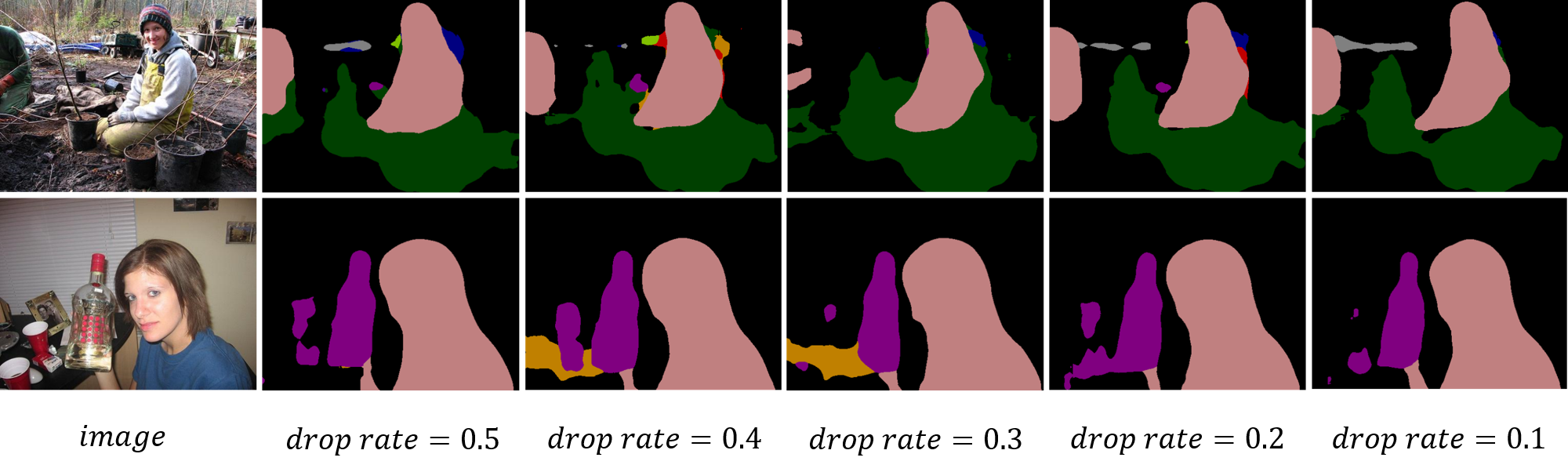}
\caption{Results of proposed method on \emph{scribble-drop} dataset with different drop rate.}
\label{fig:dropout_result}
\end{figure}

\begin{figure}[htbp]
\centering
\includegraphics[width=1\linewidth]{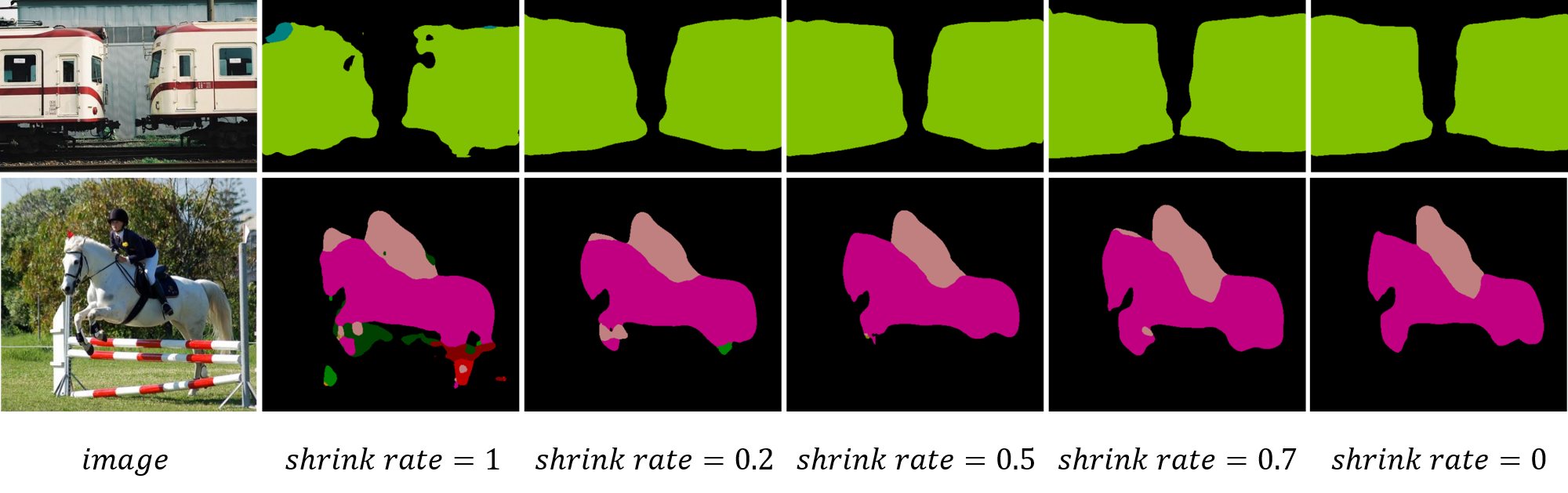}
\caption{Results of proposed method on \emph{scribble-shrink} dataset with different shrink rate.}
\label{fig:shrink_result}
\end{figure}

It should be noted that some methods, such as~\citealp{lin2016scribblesup,vernaza2017learning}, require every existing object in the image labeled. However, ours does not have this limit. To evaluate how the performance affected when the scribble randomly dropped and shrank, we further prepare two datasets, \emph{scribble-drop} and \emph{scribble-shrink} modified from \emph{scribblesup} as described before. We conduct experiments with different drop and shrink rate and show the mIoU scores in Fig.~\ref{fig:datasets}. Besides, in Tab.~\ref{tab:drop_ratios}, we also show the performance drop ratios under different drop and shrink rates. The methods used are identical to the ones in Tab.~\ref{tab:ablation}.
It can be seen that the proposed method demonstrates certain robustness when the drop rate and shrink rate increase, even all the scribbles were shrunk to points (point-supervised). 

\subsection{Qualitative Results}
We show visual comparison in Fig.~\ref{fig:VisualizationResults},  Fig.~\ref{fig:dropout_result} and  Fig.~\ref{fig:shrink_result}. Fig.~\ref{fig:VisualizationResults} presents results of $NCL$, $KCL$ and ours on \emph{scribblesup} dataset. With the proposed random walk (RW) and self-supervision on eigenspace (SS), the results are gradually refined, and the complete method outperforms others clearly and shows significant promotion over the baseline.
Fig.~\ref{fig:dropout_result} and Fig.~\ref{fig:shrink_result} further demonstrate our results on \emph{scribble-drop} and \emph{scribble-shrink}. Here no CRF using for better evaluation. It can be seen some details are missing when the annotations for training are gradually shrunk, but the main parts are preserved well. As for the random drop, our method shows promising robustness. When each scribble was dropped with 50\% probability during training, the prediction does not degrade much.
(The results and scores in this section are all from the validation set. Please check supplementary for more results.)

\subsection{Efficiency}
We analyze the efficiency after adding random walk and self-supervision in Tab.~\ref{tab:efficiency}. 
Since the self-supervision branch shares the same trainable-parameters with the main branch, no extra parameters are needed. Besides, the self-supervision output will act as the target, so no intermediate features need to store. Finally, self-supervision is only conducted during training. Consequently, in our implementation, the time and memory cost by random walk and self-supervision is acceptable.

\begin{table}
    \centering
    \setlength{\belowcaptionskip}{3pt}
    \caption{Trainable-parameters (P), memory (M), and inference speed (S) statistics when only using baseline and changes after gradually adding random walk and self-supervision, respectively.}
    \begin{tabular}{l|c|c|c}
    \toprule[1pt]
    &P&M&S\\
    \hline
    \hline
    Baseline&23.61 M&1090.82 MB&5.87 it/s\\
    
    +Random Walk&+0.31 M&+9.3 MB&-0.07 it/s\\
    
    +Self-supervision&+0 M&+2.7 MB&-0 it/s\\
    \bottomrule[1pt]
    \end{tabular}
    \label{tab:efficiency}
\end{table}

\section{Conclusion}
In this work, we present a scribble-supervised semantic segmentation method. The insight is to guide the network to produce uniform and consistent predictions by embedding a random walk process on the neural representation and imposing a self-supervision on the neural eigenspace. Thoroughly ablation studies and intermediate visualization have verified the effectiveness of proposed components. Finally, the complete method reaches state-of-the-art performance compared with others, even is comparable to the full-label supervised ones. Moreover, the proposed method shows robustness to the data with randomly dropped or shrank labels. 

\bibliographystyle{IEEEtran}

\end{document}